\documentclass[10pt,twocolumn,letterpaper]{article}

\usepackage{cvpr}              

\definecolor{cvprblue}{rgb}{0.21,0.49,0.74}
\usepackage[pagebackref,breaklinks,colorlinks,allcolors=cvprblue]{hyperref}
\usepackage[utf8]{inputenc} 
\usepackage[T1]{fontenc}    
\usepackage{url}            
\usepackage{booktabs}       
\usepackage{amsfonts}       
\usepackage{nicefrac}       
\usepackage{microtype}      
\usepackage{xcolor}       
\usepackage{xkcdcolors}
\usepackage{times}
\usepackage{epsfig}
\usepackage{graphicx}
\usepackage{placeins}
\usepackage{multirow}
\usepackage[skip=5pt]{caption}
\usepackage{rotating}
\usepackage{cancel}
\usepackage{setspace}

\usepackage{bm}
\usepackage{bibunits} 

\usepackage{enumitem}
\usepackage{amssymb,amsthm}
\usepackage{amsmath}
\usepackage{graphicx}
\usepackage{wrapfig,lipsum,booktabs}

\usepackage{epsfig}
\usepackage{tikz}
\usetikzlibrary{spy}
\usepackage{algpseudocode}
\usepackage{algorithm}
\usepackage{mathrsfs}

\usepackage{nicefrac}       
\usepackage{booktabs}       

\usepackage{thmtools,thm-restate}
\usepackage{listings}
\usepackage{lstautogobble}  %
\usepackage{color}          %
\usepackage{zi4}            %
\definecolor{bluekeywords}{rgb}{0.13, 0.13, 1}
\definecolor{greencomments}{rgb}{0, 0.5, 0}
\definecolor{redstrings}{rgb}{0.9, 0, 0}
\definecolor{graynumbers}{rgb}{0.5, 0.5, 0.5}
\usepackage{subcaption}        
\usepackage{siunitx}               
\usepackage[export]{adjustbox}
\usepackage{makecell}
\usepackage{url}
\usepackage{tikz, pgfplots}
\usetikzlibrary{positioning}
\usepackage{capt-of}
\usepackage{diagbox}

\def\eqref#1{(\ref{#1})}

\usepackage{colortbl}

\usepackage{mdframed}
\usepackage{transparent}

\usepackage{amsmath,amsfonts,bm}

\def\eqref#1{(\ref{#1})}

\def\1{\bm{1}}

\DeclareMathAlphabet{\mathsfit}{\encodingdefault}{\sfdefault}{m}{sl}
\SetMathAlphabet{\mathsfit}{bold}{\encodingdefault}{\sfdefault}{bx}{n}

\newcommand{\R}{\mathbb{R}}

\newcommand{\xb}{{\boldsymbol x}}

\newcommand{\wb}{{\boldsymbol w}}

\newcommand{\x}{{\boldsymbol x}}

\newcommand{\epsilonb}{{\boldsymbol \epsilon}}

\newcommand{\Ed}{{\mathbb E}}
\newcommand{\Rd}{{\mathbb R}}

\newcommand{\Nc}{{\mathcal N}}
\newcommand{\Mc}{{\mathcal M}}

\newcommand\norm[1]{\left\lVert#1\right\rVert}

\newcommand{\Eb}{{\mathbb E}}

\newcommand{\alphabar}{{\bar \alpha}}

\usepackage{amsthm}

\renewcommand{\c}{{\boldsymbol c}}

\usetikzlibrary{positioning}
\usepackage{capt-of}
\usepackage{diagbox}

\definecolor{C0}{rgb}{0.121569, 0.466667, 0.705882}
\definecolor{C1}{rgb}{1.000000, 0.498039, 0.054902}
\definecolor{C2}{rgb}{0.172549, 0.627451, 0.172549}
\definecolor{C3}{rgb}{0.839216, 0.152941, 0.156863}
\definecolor{C4}{rgb}{0.580392, 0.403922, 0.741176}
\definecolor{C5}{rgb}{0.549020, 0.337255, 0.294118}
\definecolor{C6}{rgb}{0.890196, 0.466667, 0.760784}
\definecolor{C7}{rgb}{0.498039, 0.498039, 0.498039}
\definecolor{C8}{rgb}{0.737255, 0.741176, 0.133333}
\definecolor{C9}{rgb}{0.090196, 0.745098, 0.811765}
\definecolor{trolleygrey}{rgb}{0.5, 0.5, 0.5}

\definecolor{BrickRed}{rgb}{0.6,0,0}
\definecolor{RoyalBlue}{rgb}{0,0,0.8}
\definecolor{Tdgreen}{rgb}{0,0.4,0.7}
\definecolor{pinegreen}{rgb}{0.0, 0.47, 0.44}
\definecolor{cornellred}{rgb}{0.7, 0.11, 0.11}
\definecolor{cadmiumgreen}{rgb}{0.0, 0.42, 0.24}
\definecolor{spirodiscoball}{rgb}{0.06, 0.75, 0.99}
\definecolor{mylightblue}{rgb}{0.85, 0.90, 0.94}
\definecolor{maroon}{cmyk}{0,0.87,0.68,0.32}

\usepackage{pifont}
\algnewcommand{\LineComment}[1]{\State \(\triangleright\) #1}

\definecolor{cfg}{rgb}{0.906, 0.435, 0.318}
\definecolor{cfgpp}{rgb}{0.165, 0.616, 0.561}
\definecolor{cfgnull}{rgb}{0.208, 0.565, 0.953}

\definecolor{pp}{rgb}{0.165, 0.616, 0.561}
\definecolor{teacher}{rgb}{0.906, 0.435, 0.318}
\definecolor{student}{rgb}{0.208, 0.565, 0.953}
\definecolor{ppgreen}{rgb}{0.93, 0.98, 0.9}

\newcommand{\epsstudent}{\textcolor{student}{\epsilonb_\theta}}
\newcommand{\epsteacher}{\textcolor{teacher}{\epsilonb_\psi}}
\newcommand{\epshatstudent}{\textcolor{student}{\hat{\epsilonb}_\theta^w}}
\newcommand{\epshatteacher}{\textcolor{teacher}{\hat{\epsilonb}_\psi^w}}

\newcommand{\tweediestudent}{\textcolor{student}{\hat\x_0^\theta}}
\newcommand{\tweedieteacher}{\textcolor{teacher}{\hat\x_0^\psi}}

\newcommand{\tweediecstudent}{\textcolor{student}{\hat{\x}_\c^\theta}}
\newcommand{\tweediecteacher}{\textcolor{teacher}{\hat{\x}_\c^\psi}}

\newcommand{\tweediepp}{\textcolor{pp}{\hat{\x}_{\text{new}}^\theta}}
\newcommand{\tweedieppc}{\textcolor{pp}{\hat{\x}_{\text{new}, \c}^\theta}}
\newcommand\ccw[1]{\cellcolor[rgb]{0.93, 0.98, 0.9}{#1}}  

\def\paperID{15140} 
\def\confName{CVPR}
\def\confYear{2025}

\title{Inference-Time Diffusion Model Distillation}

\author{
    Geon Yeong Park$^{1}$ \qquad Sang Wan Lee$^{1,3*}$ \qquad Jong Chul Ye$^{1,2*}$ \\
$^{1}$Bio and Brain Engineering, $^{2}$Kim Jaechul Graduate School of AI, $^{3}$Brain and Cognitive Sciences \\
Korea Advanced Institute of Science and Technology (KAIST), Daejeon, Korea \\
{\tt\small \{pky3436, sangwan, jong.ye\}@kaist.ac.kr}
}

\begin{document}

\twocolumn[{%
\renewcommand\twocolumn[1][]{#1}%
\maketitle
\vspace{-1cm}
\begin{center}
    \centering
    \captionsetup{type=figure}
    \includegraphics[width=0.95\textwidth]{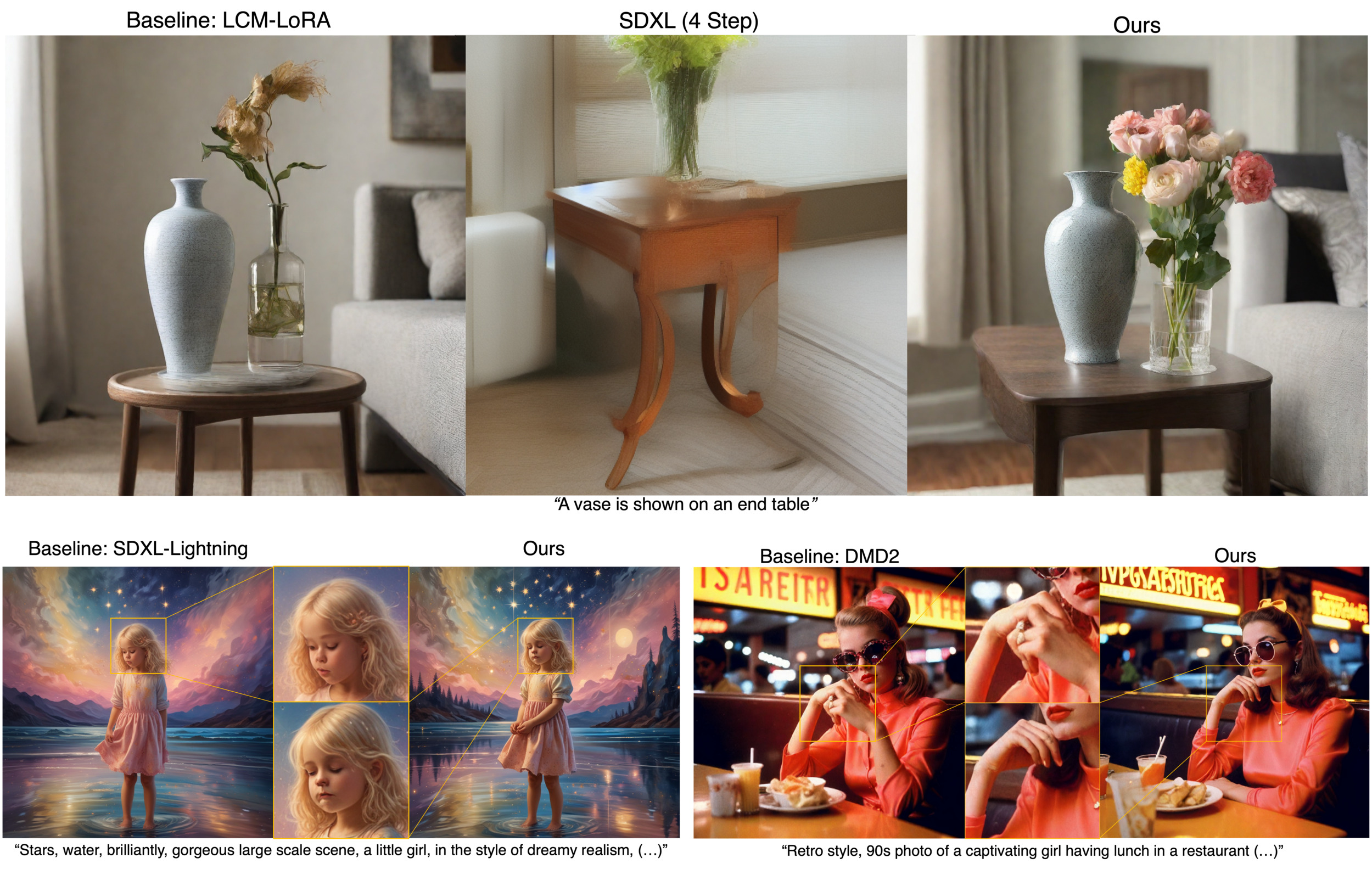}
    \captionof{figure}{{Experiments comparing baselines (LCM-LoRA \cite{luo2023lcmlora}, SDXL-Lightning \cite{lin2024sdxl}, DMD2 \cite{yin2024improved}), teacher model (SDXL \cite{podell2023sdxl}), and the proposed framework. Results  are generated with 4 step sampling. We improve the visual fidelity and textual alignment of student baselines by conducting a \textit{inference-time} diffusion distillation with the guidance of teacher model, e.g. SDXL, in early sampling stages (first 1 step).
    }
    }
    \vspace{-0.05cm}
    \label{fig: teaser}
\end{center}%
}]


\begin{abstract}
Diffusion distillation models effectively accelerate reverse sampling by compressing the process into fewer steps. However, these models still exhibit a performance gap compared to their pre-trained diffusion model counterparts, exacerbated by distribution shifts and accumulated errors during multi-step sampling. To address this, we introduce Distillation++, a novel inference-time distillation framework that reduces this gap by incorporating teacher-guided refinement during sampling. Inspired by recent advances in conditional sampling, our approach recasts student model sampling as a proximal optimization problem with a score distillation sampling loss (SDS). To this end, we integrate distillation optimization during reverse sampling, which can be viewed as teacher guidance that drives student sampling trajectory towards the clean manifold using pre-trained diffusion models. Thus, Distillation++ improves the denoising process in real-time without additional source data or fine-tuning. Distillation++ demonstrates substantial improvements over state-of-the-art distillation baselines, particularly in early sampling stages, positioning itself as a robust guided sampling process crafted for diffusion distillation models. Code: \href{https://github.com/geonyeong-park/inference_distillation}{here}.
\end{abstract}

\section{Introduction}

Diffusion models have significantly advanced image generation by producing high-quality samples through an iterative refinement process that gradually denoises an initial noise vector. This refinement can be viewed as solving the reverse generative Stochastic Differential Equation (SDE) or Ordinary Differential Equations (ODE), a counterpart to a prescribed forward SDE/ODE. 

Despite achieving unprecedented realism and diversity, diffusion models face a critical challenge: slow sampling speed. The progressive denoising process is computationally expensive because solving the reverse SDE/ODE requires fine discretization of time steps to minimize discretization errors considering the curvature of the diffusion sampling trajectory \citep{karras2022elucidating}. This leads to an increased number of function evaluations (NFE), where typical diffusion sampling necessitates tens to hundreds of NFE, limiting its use in user-interactive creative tools.

To address these limitations, various works have proposed to accelerate diffusion sampling. One promising avenue is distillation models, which distill the pre-trained diffusion models (teacher model) by directly estimating the integral along the Probability Flow ODE (PF-ODE) trajectory \citep{song2020score}. This effectively amortizes the computational cost of sampling into the training phase. Recent advances in distillation methods have led to the emergence of a one-step image generator; however, few-step distillation models (student models) are also often preferred in terms of image quality.

Despite progress, distillation models still face challenges, particularly in bridging the performance gap between the teacher model and its distilled student counterpart. The primary issues include potential suffer from accumulated errors in multi-step sampling or iterative training. For example, \cite{kim2023consistency} identifies potential issues of multi-step sampling with models estimating the zero-time endpoint of PF-ODE. As an empirical demonstration, they show that the generation quality of consistency models does not improve as NFE increases. Similarly, \cite{geng2024consistency} shows that consistency errors can accumulate across time intervals, leading to unstable training. \cite{yin2024improved, garrepalli2024ddil} warn the general training/inference mismatch observed in the multi-step sampling of student models. While prior works aim to mitigate this gap by introducing real training datasets \cite{yin2024improved, sauer2023adversarial}, it may face a potential distribution shift between datasets of teacher and student, leading to suboptimal performance on out-of-distribution (OOD) prompts. 

In this work, we aim to overcome this fundamental gap between teacher and student by proposing a novel {\em inference-time} distillation framework called  \textit{Distillation++}. Distillation++ is a novel symbiotic distillation framework that distills the teacher model to a student model {throughout the \textit{sampling} process, in contrast to prior works which distill only during \textit{training} process}. In particular, inspired by the recent advances in text-conditional sampling \citep{kim2024dreamsampler, chung2024cfg++, chung2023decomposed}, we first recast the diffusion sampling of student models to the proximal optimization problem and regularize its sampling path with a score distillation sampling loss (SDS, \cite{poole2022dreamfusion}) which opens an important opportunity for external guidance. Based on this insight, we develop a teacher-guided sampling process that inherently minimizes the SDS loss by leveraging the pre-trained diffusion model as a teacher evaluating student's denoised estimates during sampling. Specifically, intermediate estimates of student models are refined towards the clean manifold by minimizing the SDS loss, computed 
using the teacher model. 

This inference-time distillation fosters a life-long partnership between teacher and student, allowing continuous teacher guidance beyond the training phase. Our empirical results demonstrate that this data-free distillation approach significantly improves student model performance, particularly in the early stages of sampling with minimal computational costs. 
We believe this approach introduces a new opportunity for inference-time distillation, a concept that has not been previously explored.

Our contributions can be summarized as follows:
\begin{itemize}[noitemsep]
    \item We introduce Distillation++, a novel inference-time distillation framework where teacher guides sampling process so that it closes the gap between student and teacher during sampling with affordable computational costs. 
    \item The proposed framework is generally compatible with various student models, ranging from ones directly predicting the PF-ODE endpoint \citep{song2023consistency, luo2023latent, yin2024one} to the progressive distillation branches \citep{salimans2022progressive, lin2024sdxl}. We also demonstrate its general applicability with various solvers, including Euler and DPM++ 2S Ancestral \citep{lu2022dpmpp}.
    \item To the best of our knowledge, Distillation++ is a first tuning-free and data-free inference-time distillation framework that serves as a viable post-training option for improving distillation model sampling.
\end{itemize}

\begin{figure*}[t]
    \centering
    \includegraphics[width=\textwidth]{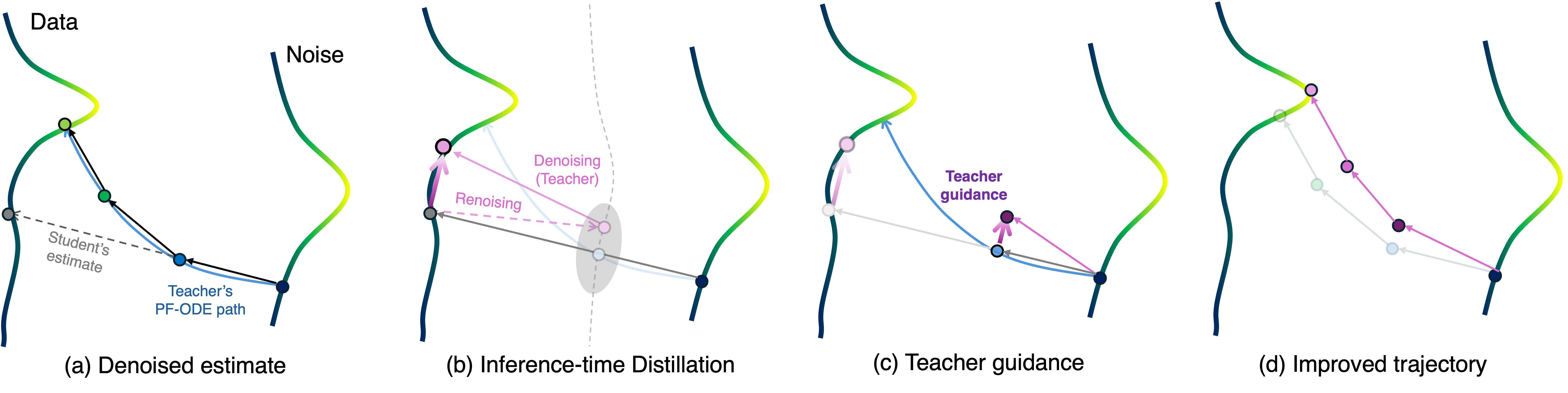}
    \caption{{\textbf{Overview.} (\textbf{a}) Diffusion models (in blue) sample by solving the PF-ODE, requiring a computationally expensive integral from time $T$ to 0. Student models (in black) accelerate sampling by approximating this integral, but their (initial) estimates are often suboptimal. (\textbf{b}) To bridge this gap post-training, we propose an \textit{inference-time} distillation. Specifically, we refine the student models' initial estimates by refining them towards teacher estimates, obtained by consecutive renoising and denoising, as in \eqref{eq: teacher guidance2}. (\textbf{c}) This process functions as a form of teacher guidance in \eqref{eq: teacher guidance3}, steering the sampling trajectory closer to the teacher model's distribution, thereby (\textbf{d}) improving the sampling path.}
    }
    \label{fig:main}
\end{figure*}

\section{Background}
\noindent \textbf{Diffusion models.} Diffusion models aim to generate samples by learning the reversal of a prescribed diffusion forward process. In discrete setting with a total of $N$ noise scales, define the fixed forward diffusion kernel as follows:
\begin{align}
\label{eq: diffusion kernel}
p(\xb_t | \xb_{t-1}) &= \Nc (\xb_t | \beta_t \xb_{t-1}, (1-\beta_t) I), \\ 
p_t(\xb_t | \xb_0) &= \Nc (\xb_t | \sqrt{\alphabar} \xb_0, (1-\alphabar) I),
\end{align}
where $\x_0 \in \mathbb{R}^d \sim p_0(\x)$ is given as a clean sample, $\beta_t$ denotes a noise schedule discretized from $\beta(t): \mathbb{R} \rightarrow \mathbb{R} > 0$, $\alpha_t := 1 - \beta_t$ and $\alphabar_t := \Pi_{i=1}^t \alpha_i$. Then as $N \rightarrow \infty$, the underlying forward noising process can be expressed as the forward It\^{o} SDE \citep{song2021scorebased} given $\x(t) \in \mathbb{R}^d$:
\begin{align}
\label{eq: forward sde}
d\x = - \frac{\beta(t)}{2} \x dt + \sqrt{\beta(t)} d\wb,
\end{align}
where $\wb$ is the $d$-dimensional standard Wiener process. Then the counterpart reverse SDE is defined as follows \citep{anderson1982reverse}:
\begin{align*}
d\x = \Big[-\frac{\beta(t)}{2}\x - \beta(t) \nabla_{\x} \log p_t(\x) \Big] dt + \sqrt{\beta(t)} d \bar{\wb}, 
\end{align*}
where $\bar{\wb}$ is the $d$-dimensional standard backward Wiener process. The deterministic counterpart PF-ODE \citep{song2021scorebased} can be similarly defined.
Then the goal of diffusion model training is to approximate a score function $\nabla_{\x} \log p_t(\x)$ by denoising score matching (DSM):
\begin{align*}
    \min_{\theta} \Eb_{\xb(t), \xb_0, t} \big[ \norm{ \boldsymbol{s}_{\theta} (\xb(t), t) - \nabla_{\xb(t)} \log p_t(\xb(t) | \xb_0) } \big],
\end{align*}
where $\x_0 \sim p_0(\x)$ denotes a clean sample. It can be shown that this score matching is equivalent to the epsilon matching with different parameterization of residual denoiser $\epsilonb_\theta$: 
\begin{align}
    \label{eq: epsilon matching}
    \min_{\theta} \Eb_{\xb(t), \xb_0, \epsilonb \sim \Nc(0, I)} \big[ \norm{\epsilonb_{\theta} (\xb(t), t) - \epsilonb} \big],
\end{align}
and $\boldsymbol{s}_{\theta^*}(\xb(t), t) \simeq - \frac{\xb(t) - \sqrt{\alphabar_t} \xb_0 }{1-\alphabar} = - \frac{1}{\sqrt{1-\alphabar_t}} \epsilonb_{\theta^*} (\xb(t), t)$.

Based on this, generative sampling can be performed by solving the PF-ODE, equivalent to computing the integral from time $T$ to $0$, approximated by various off-the-shelf ODE solvers. For instance, a single iteration of DDIM sampling \citep{song2020denoising} reads:
\begin{align}
\label{eq: ddim_update}
    \hat{\xb}_0^\theta(t) &= (\x_t - \sqrt{1-\bar\alpha_t} \epsilonb_\theta(\x_t, t))/\sqrt{\bar\alpha_t} \\
    \x_{t-1} &= \sqrt{\bar\alpha_{t-1}} \hat{\xb}_0^\theta(t) + \sqrt{1-\bar\alpha_{t-1}} \epsilonb_\theta(\x_t, t)),
\end{align}
where  $\xb_t \in \Mc_t$ with noisy manifold $\Mc_t$, and $\hat{\x}_0(t) =  \Ed[\x_0|\x_t]$ is the denoised estimate, which can be equivalently derived using Tweedie's formula \citep{efron2011tweedie}. 

For a text-conditional sampling with text embedding $\c$, the classifier-free guidance (CFG, \cite{ho2022classifier}) is widely leveraged:
\begin{align}
\hat{\epsilonb}_\theta^w(\x_t, t, \c) = \epsilonb_\theta(\x_t, t, \varnothing) + w(\epsilonb_\theta(\x_t, t, \c) - \epsilonb_\theta(\x_t, t, \varnothing)), \nonumber
\end{align}
where $\c=\varnothing$ refers to the null text embedding. For notational simplicity, we will interchangably use $\epsilonb_\theta(\x_t, t, \varnothing), \epsilonb_\theta(\x_t, \c), \epsilonb_\theta(\x_t)$ and similarly $\epsilonb_\theta(\x_t, t, \c), \epsilonb_\theta(\x_t, \c)$ unless ambiguity arises.


\noindent \textbf{Diffusion acceleration.} While diffusion models generate high-quality samples with a relatively stable training procedure of (\ref{eq: epsilon matching}), the numerical integration of the PF-ODE is computationally expensive. Two primary strategies have been developed to accelerate this process. The first strategy corresponds to the fast diffusion samplers which improves time-discretized numerical integration methods \cite{song2020denoising, lu2022dpm, jolicoeur2021gotta, karras2022elucidating}. Despite its interesting progress, further reducing NFE can significantly degrade performance in practice. 

Alternatively, diffusion distillation has emerged as a promising approach to accelerate the sampling by indirectly estimating the entire integral of PF-ODE. For instance, \cite{luhman2021knowledge} trains the denoising student that predicts the endpoint of the PF-ODE given an initial noise vector. Similarly, \cite{yin2024one} trains a one-step generator by matching the pre-compute diffusion outputs with using the distribution matching training objective. \cite{song2023consistency, berthelot2023tract, gu2023boot} further learn to map a different point on the ODE trajectory to its endpoint or to the boundaries of sub-intervals. In a related yet distinct approach, progressive distillation methods have been developed considering the expensive computational cost of simulating the full denoising trajectory for each loss function evaluation. These methods often iteratively train a series of student models, each halving the number of sampling steps required by the previous model.

\section{Main Contribution: Distillation++}
\vspace{-0.2cm}
\label{sec: distillation++}
While both improved sampling and distillation methods have made significant progress in addressing the speed-quality trade-off, there has been limited advances in integrating these methods through the design of specialized solvers for improving multi-step distillation models. This can be attributed to two primary reasons. First, off-the-shelf ODE solvers typically require the estimation of the tangent gradient direction along the solution trajectory. In contrast, many distillation models \citep{song2023consistency, luo2023latent, yin2024one, yin2024improved} directly predict the endpoint of the trajectory on the side of data distribution at any given time point, avoiding the need for direct trajectory estimation. Consequently, the multi-step sampling procedures of these student models are reduced to simple iterative processes involving random noise injection and subsequent denoising steps \citep{song2023consistency}. Furthermore, distillation models often sample only 2-8 steps, which may restrict the design space for solvers, including higher-order ones. 

These constraints limit post-training options for improving distillation model sampling, despite the performance gap between multi-step student and teacher models. To address this, 
Distillation++ leverages large-scale pre-trained diffusion models as a teacher signal during the early-stage sampling process (e.g. first 1-2 steps), which substantially improves the overall sampling trajectory as shown in Fig. \ref{fig:main}.  More details follow.

\subsection{Derivations of Distillation++}
\vspace{-0.2cm}
\label{sec: derivation}
For a better multi-step sampling of student models, we derive a novel data-free distillation process by integrating the teacher guidance as an optimization problem within the reverse sampling process. Let $\textcolor{student}{\theta} \in \Rd^D$ and $\textcolor{teacher}{\psi} \in \Rd^D$ parameterize the residual denoiser of student and teacher diffusion models, respectively. Then, our objective is to define a guidance loss function $\ell_{\text{distill}}$ that, when minimized under the symbiotic guidance of the teacher model, progressively aligns the \textit{student's} intermediate estimates $\tweediestudent(t)$ with the teacher model's distribution. We present such guidance loss function $\ell_{\text{distill}}$ as a score distillation sampling loss (SDS) with respect to the \textit{teacher model} ($\psi$) as follow:
\begin{align}
\label{eq: sds loss}
\ell_{\text{distill}}(\x; \psi, s) &= \norm{ \epsteacher(\underbrace{\sqrt{\alphabar_s} \x + \sqrt{1-\alphabar_s} \epsilonb}_{\xb_s}, \varnothing) - \epsilonb}_2^2 \nonumber \\
&= \norm{ \frac{\xb_s - \sqrt{\alphabar_s} \tweedieteacher(s)}{\sqrt{1-\alphabar_s}} -\frac{\xb_s - \sqrt{\alphabar_s} \x}{\sqrt{1-\alphabar_s}}}_2^2 \nonumber \\
&= \frac{\alphabar_s}{1 - \alphabar_s} \norm{\x - \tweedieteacher(s)}_2^2, 
\end{align}
where $\xb_s = \sqrt{\alphabar_s} \x + \sqrt{1-\alphabar_s} \epsilonb$ with a perturbation time $s>0$, $\epsilonb \sim \Nc(\epsilonb | 0, I), \x \in \Mc$ with a clean data manifold $\Mc$, and $\tweedieteacher(s)$ follows the Tweedie's formula in (\ref{eq: ddim_update}) which is denoised by the teacher model $\psi$. For simplicity, we consider the text-unconditional version with the null-text embedding $\varnothing \in \R^{d'}$ in this subsection. This loss represents an ideal condition that high-quality denoised estimates should satisfy: the ideal student samples should be well reconstructed from random perturbations followed by denoising using large-scale pre-trained teacher diffusion models. Variants of the SDS framework \citep{nguyen2024swiftbrush, luo2024diff, yin2024one} have thus frequently been employed as key components in recent diffusion distillation training procedure.

Then, we can now integrate the optimization step of $\ell_{\text{distill}}$ in terms of denoised student estimates $\tweediestudent(t)$, resulting DDIM sampling process (\ref{eq: ddim_update}).
A potential concern includes the feasibility of gradient descent due to the intractable Jacobian computation: $\frac{d\ell}{d\x} = \frac{d\ell}{d\epsteacher} \frac{d\epsteacher}{d\x}$. To circumvent this,
 by following the prior work on Decomposed Diffusion Sampling (DDS) \citep{chung2023decomposed}, which bypasses direct computation of the score Jacobian, we have
\begin{align}
\label{eq: teacher guidance1}
    \x_{t-1} &= \sqrt{\alphabar_{t-1}} \Big( \tweediestudent(t) - \gamma_t \nabla_{\tweediestudent(t)} \ell_{\text{distill}} \big(\tweediestudent(t); \psi, s \big)\Big) \nonumber\\
    &+ \sqrt{1-\bar\alpha_{t-1}} \epsstudent(\x_t, t),
\end{align}
where $\gamma_t > 0$ refers to the step size.  \cite{chung2023diffusion, chung2024cfg++} supports that this update allows precise transition to the subsequent noisy manifold $\Mc_{t-1}$ under some manifold assumption.
This gives us a simple single DDIM sampling iterate as follows:
\begin{align}
\label{eq: teacher guidance2}
    \tweediepp(t) &= \tweediestudent(t) - \lambda \big( \tweediestudent(t) - \tweedieteacher(s) \big) \nonumber \\
    &= (1-\lambda) \tweediestudent(t) + \lambda \tweedieteacher(s), \\
    \x_{t-1} &= \sqrt{\alphabar_{t-1}} \tweediepp(t) + \sqrt{1-\bar\alpha_{t-1}} \epsstudent(\x_t, t) \nonumber,
\end{align}
where $\lambda = \frac{2 \gamma_t \sqrt{\alphabar_t}}{\sqrt{1 - \alphabar_t}}$. Note that the updated estimate $\tweediepp(t)$ can be obtained by the interpolation between initial student estimate $\tweediestudent(t)$ and the revised teacher estimate $\tweedieteacher(s)$ which is the denoised estimated from different time index $s\neq t$. 
 Thus, the goal is to update $\tweediestudent(t)$ towards the clean manifold well-aligned with the teacher model distribution. 


\noindent\textbf{Renoising strategy.} Note that \eqref{eq: teacher guidance2} is not a simple interpolation between the student and teacher during the sampling.
Indeed, the time step schedule of $s\neq t$ plays a crucial role in performance. 
Considering that the teacher model is well-trained on fine level of timesteps, it is compatible with a broad range of renoising timestep $s$. That said, as our approach recasts distillation more as a guided sampling rather than a mere training process, we adopt a decreasing time step schedule for $s$ as $s=t-1$ following the sampling process (Fig. \ref{fig:main}b), reminiscent of schedules used in \citep{kim2024dreamsampler, zhu2023hifa, wu2024consistent3d}. This is in contrast to conventional random timestep $s$ scheduling \cite{poole2022dreamfusion}. Intuitively, as the student model often learns to leap towards the end-point of each sub-interval (Fig. \ref{fig:main}a), refining this large-step update direction at the terminal of each sub-interval with teacher model may better guide the sampling trajectory (Fig. \ref{fig:main}c,d). Empirical evidences and more discussions are provided in Table \ref{table: time schedule}.

{\noindent\textbf{Teacher guidance.}  For a better understanding of Distillation++, we provide reparameterization of (\ref{eq: teacher guidance2}), by intentionally assuming $\alphabar_t \approx \alphabar_s$. This leads to the formulation of \textit{teacher guidance}, drawing parallels to the CFG denoising mechanism:}
\begin{align}
\label{eq: teacher guidance3}
    \tweediepp(t) &= \frac{\tilde{\x}_t - \sqrt{1-\alphabar_t} \big( \epsstudent(\x_t, t) + \lambda \big( \epsteacher(\x_s, s) - \epsstudent(\xb_t, t) \big) }{\sqrt{\alphabar_t}} \nonumber \\
    \x_{t-1} &= \sqrt{\alphabar_{t-1}} \tweediepp(t) + \sqrt{1-\bar\alpha_{t-1}} \epsstudent(\x_t, t),
\end{align}
where $\lambda > 0$ serves as a guidance scale, and $\tilde{\x}_t = (1 - \lambda) \x_t + \lambda \x_s$. We conduct experiments using (\ref{eq: teacher guidance2}) in practice as (\ref{eq: teacher guidance3}) serves primarily as an approximation for intuition; however, \eqref{eq: teacher guidance2} and \eqref{eq: teacher guidance3} reproduce empirically similar generations with negligible differences, particularly with sampling steps $\geq 4$ (empirical comparisons in Fig. \ref{fig: teacher guidance}). 
This suggests that the success of teacher guidance primarily lies in the directional component $\lambda(\epsteacher(\x_s, s) - \epsstudent(\xb_t, t))$ , which steers the sampling trajectory closer to the teacher model distribution as in Fig. \ref{fig:main} (c-d).

Experiments reveal that a simple constant $\lambda$ performs effectively, despite the potential benefits of a more complex time-dependent setting. This is because teacher guidance is applied to only a few early sampling steps (e.g., 1-2 steps) for computational efficiency. This approach minimizes the additional computational costs of inference-time distillation and performs effectively, given that the student model is initialized from a pre-distilled state rather than from scratch. More discussions are provided in Sec. \ref{sec: result}.

\noindent\textbf{Remarks.}
\label{sec: remark}
Distillation++ is compatible with various distillation models, especially those directly predicting the PF-ODE endpoint, such as consistency models \citep{song2023consistency} , which are incompatible with ODE solvers like DDIM in Sec. \ref{sec: derivation}. While Sec. \ref{sec: derivation} simplifies by focusing on a specific solver, Distillation++ is generalizable, even with simple iterative processes involving denoising followed by stochastic renoising steps, as in multi-step consistency models. Starting with an initial endpoint estimate from the consistency model, we randomly perturb it with time $s$ and derive the update direction using the teacher model $\psi$, emphasizing modulation of the denoising rather than the renoising process.

\vspace{-0.1cm}
\subsection{Text-conditional sampling and other solvers}
Section \ref{sec: derivation} focused on the text-unconditional setting for simplicity. Here we incorporate the teacher guidance with widespread classifier-free guidance (CFG, \cite{ho2022classifier}) for text-conditional sampling. Let $\tweediecstudent(t)$ denote the text-conditioned clean estimate defined as:
\begin{align}
\label{eq: cfg tweedie}
    \tweediecstudent(t) = \frac{\tilde{\x}_t - \sqrt{1-\alphabar_t} \epshatstudent(\x_t, \c)}{\sqrt{\alphabar_t}},
\end{align}
where we will interchangeably use $\tweediecstudent(\x_t)$ and similarly define $\tweediecteacher(t)$. Then, based on \eqref{eq: teacher guidance2}, the resulting text-conditional version of $\tweediepp(t)$ in \eqref{eq: teacher guidance2} is derived as follows: $\tweedieppc(t) = (1-\lambda) \tweediecstudent(t) + \lambda \tweediecteacher(s)$. Here, $\tweedieppc(t)$ can be similarly derived in the form of teacher guidance as in \eqref{eq: teacher guidance3} with the same assumption on $\alphabar_t$:
\begin{align}
    \frac{\tilde{\x}_t - \sqrt{1-\alphabar_t} \big( \epshatstudent(\x_t, \c) + \lambda \big( \epshatteacher(\x_s, \c) - \epshatstudent(\xb_t, \c) \big) }{\sqrt{\alphabar_t}}. \nonumber
\end{align}
The overall pipeline is outlined in Algorithm 1 in Supplementary.


While Section \ref{sec: derivation} is delineated with DDIM or equivalently Euler solvers, the underlying principle can be shared with other conventional solvers, e.g. Karras \citep{karras2022elucidating}, PNDM \citep{liu2022pseudo}, etc. To demonstrate this extension, we consider a DPM-solver++ 2S \citep{lu2022dpmpp} with CFG++ \citep{chung2024cfg++} in VE-SDE setting \citep{song2020score}. Specifically, define $\sigma_t := e^{-t},\,h_i := t_i - t_{i-1},$ $r_i := h_{i-1}/h_{i}$ and initialize $\x_{t_0}$ with standard Gaussian noise. DPM-solver++ 2S introduces an additional intermediate time step $\{s_i\}_{i=1}^M$ with $t_i > s_{i+1} > t_{i+1}$. Let $r_i = \frac{s_i - t_{i-1}}{t_i - t_{i-1}}$. Then, an iterate of DPM-solver++ 2S with CFG++ reads:
\begin{equation}
\begin{split}
    \label{eq: dpm++}
    \bm{u}_i &= e^{-r_ih_i}\x_{t_{i-1}} + (1 - e^{-r_ih_i})\hat{\x}_\varnothing^\theta(\x_{t_{i-1}}), \\
    \x_{t_i} &= \hat{\x}_\varnothing^\theta(\x_{t_{i-1}}) - e^{-h_i}\hat{\x}_\varnothing^\theta(\x_{t_{i-1}}) \\
    &+ \frac{1 - e^{-h_i}}{2r_i}\left(
    \tweediecstudent(\bm{u}_i) - \hat{\x}_\varnothing^\theta(\x_{t_{i-1}}) \nonumber
    \right) + e^{-h_i}\x_{t_{i-1}},
\end{split}    
\vspace{-0.1cm}
\end{equation}
where $\tweediecstudent(\bm{u}_i)$ refers to the initial conditional denoised estimate with guidance scale $0 < \lambda < 1$, and the rest terms are related to the higher-order correction of the renoising process. That said, Distillation++ modulates the denoising process in (\ref{eq: dpm++}) by interpolating $\tweediecstudent(\bm{u}_i)$ with the teacher-revised estimate $\tweediecteacher(\bm{u}_i)$ as $\tweedieppc(\bm{u}_i)$:
\begin{align}
\label{eq: dpm++ teacher}
    \bm{u}_i &= e^{-r_ih_i}\x_{t_{i-1}} + (1 - e^{-r_ih_i})\hat{\x}_\varnothing^\theta(\x_{t_{i-1}}), \nonumber \\
    \x_{t_i} &= \hat{\x}_\varnothing^\theta(\x_{t_{i-1}}) - e^{-h_i}\hat{\x}_\varnothing^\theta(\x_{t_{i-1}}) + \\
    &\frac{1 - e^{-h_i}}{2r_i}\left(
    \tweedieppc(\bm{u}_i) - \hat{\x}_\varnothing^\theta(\x_{t_{i-1}})
    \right) + e^{-h_i}\x_{t_{i-1}}. \nonumber
\end{align}

This simple modification implies that Distillation++ is potentially compatible with various solvers, where the core principle is to regularize the denoising path with revised teacher's estimates. We use ancestral variant (DPM++ 2S A) of \eqref{eq: dpm++ teacher} by adding stochasticity in practice.

\begin{figure*}[htbp]
    \centering
    \includegraphics[width=\textwidth]{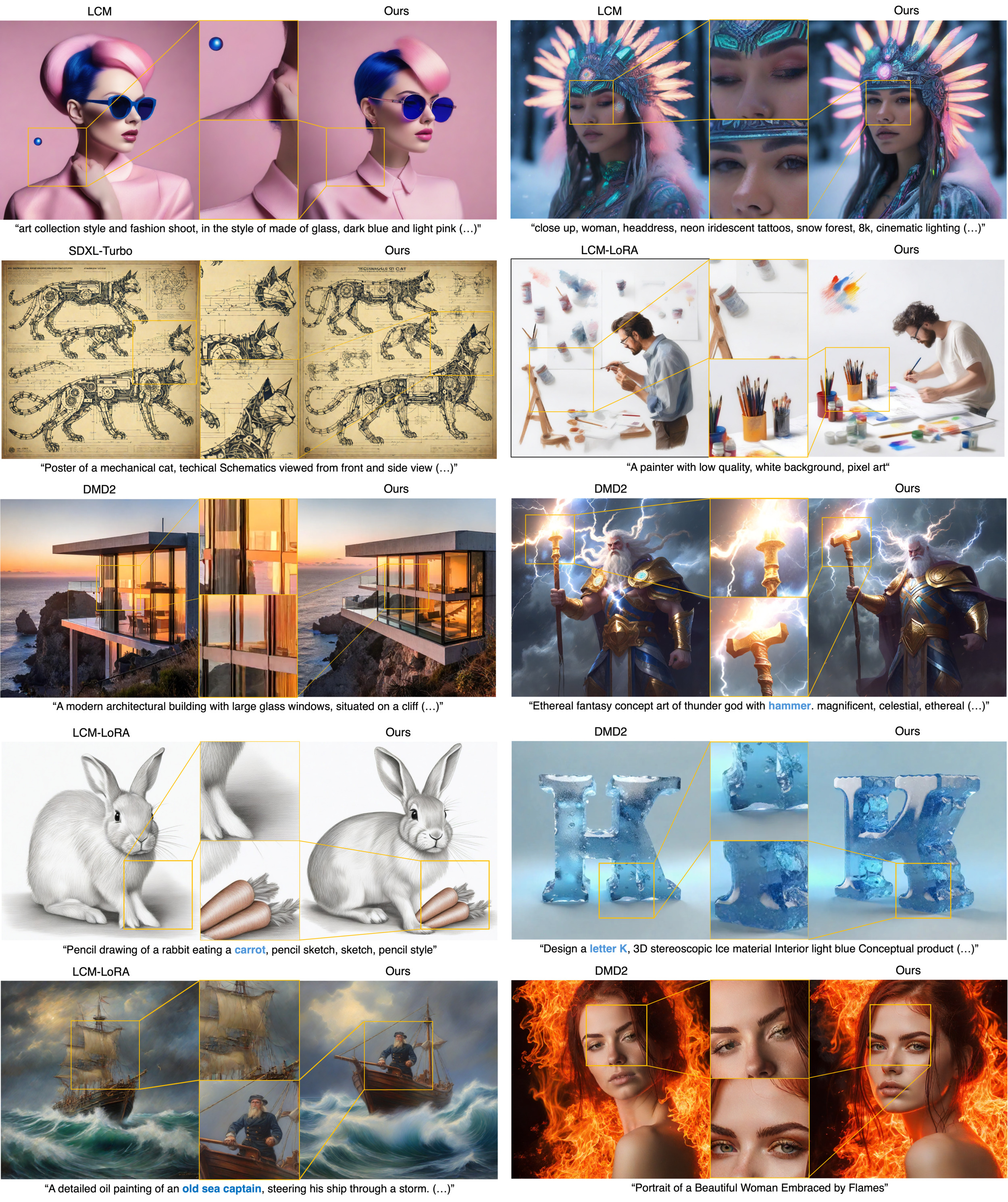}
    \caption{{Qualitative comparisons against state-of-the-art distillation baselines. Baselines using 4 sampling steps: SDXL-Lightning, DMD2, SDXL-Turbo. Baselines using 8 sampling steps: LCM, LCM-LoRA. By conducting the inferece-time distillation in early sampling stages, we reduce artifacts, improve the visual fidelity and textual alignment.}}
    \label{fig: result}
\end{figure*}

\begin{figure*}[htbp]
    \centering
    \includegraphics[width=0.95\textwidth]{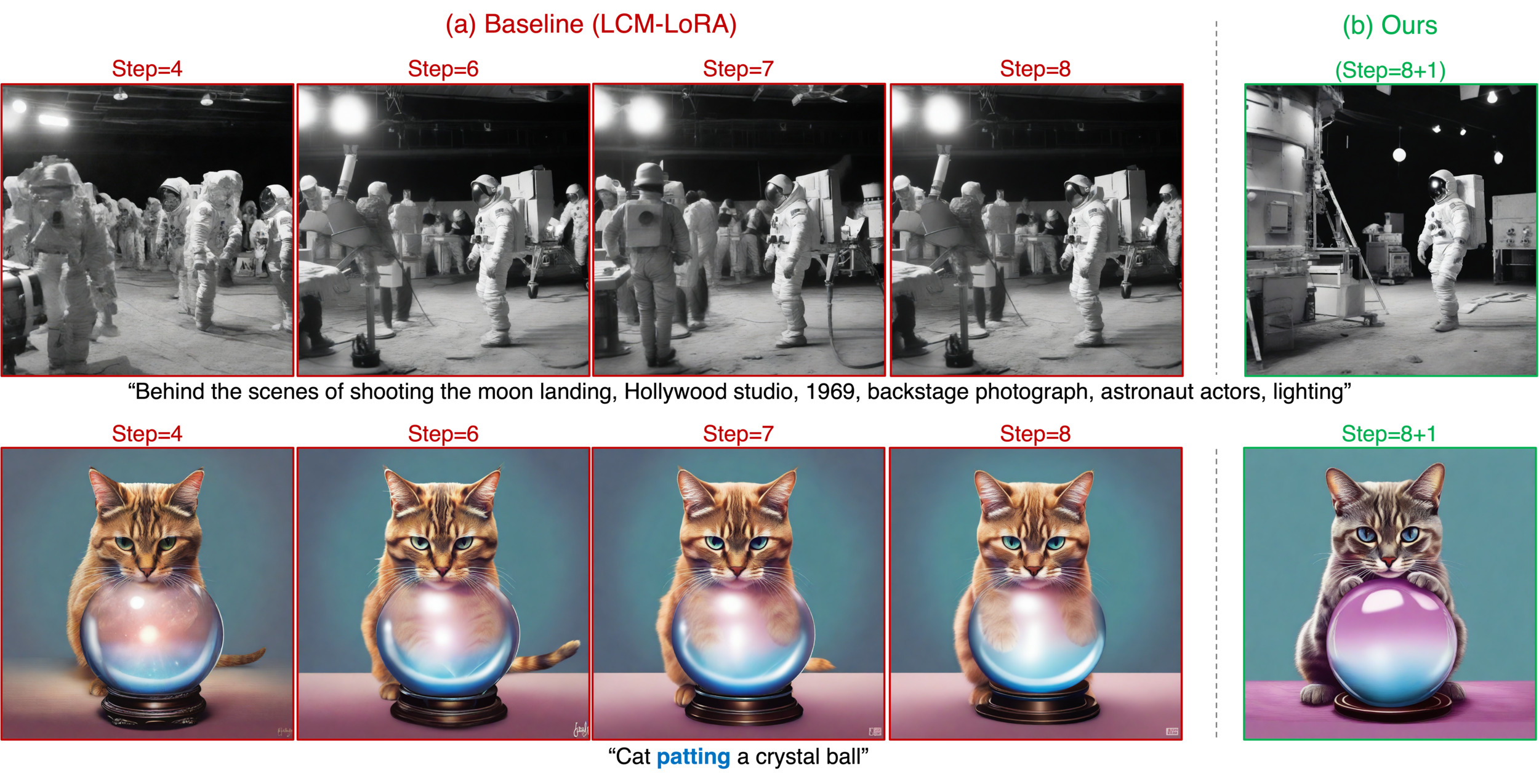}
    \caption{{(\textbf{a}) Results of baseline (LCM-LoRA) with varying number of sampling steps (4, 6, 7, 8). Increasing the number of sampling steps of student models does not guarantee improvements in textual alignment or physical feasibility. (\textbf{b}) Our improved results with inference-time distillation. Teacher guidance is applied only at the first of 8 steps (total step=8+1).}}
    \vspace{-0.3cm}
    \label{fig: cost1}
\end{figure*}

\section{Experimental results}
\subsection{Implementation details}
In our experiments, we demonstrate the impacts of Distillation++ using SDXL backbone \citep{podell2023sdxl} and its open-sourced weights. All experiments are conducted using a single NVIDIA GeForce RTX 4090. All quantitative results are obtained with one-step teacher-guided distillation at the initial sampling step of the student models (i.e., $t=T$), minimizing additional computational costs, though more frequent guidance could further enhance quality. Detailed implementation settings are provided in the appendix and the following.


\noindent\textbf{Baselines.} Distillation++ can be applied to various state-of-the-art T2I distillation baselines which are widely used with open-sourced weights. \textsf{LCM} \citep{luo2023latent} is a consistency model that operates in the image latent space of a pretrained autoencoder, solving an augmented PF-ODE in this space. \textsf{LCM-LoRA} \citep{luo2023lcmlora} extends LCM by distilling pre-trained latent diffusion models into LoRA parameters, significantly reducing memory requirements while maintaining high generation quality. \textsf{DMD2} \citep{yin2024improved} improves distribution-matching distillation (DMD, \cite{yin2024one}) by enabling multi-step sampling and eliminating the needs for expensive regression loss. \textsf{SDXL-Turbo} \citep{sauer2023adversarial} integrates both adversarial loss and score-distillation sampling loss. Building on similar adversarial training regime, \textsf{SDXL-Lightning} \cite{lin2024sdxl} inherits progressive distillation \citep{salimans2022progressive}, relaxing mode coverage and also includes a LoRA variant (\textsf{SDXL-Lightning LoRA}).

\noindent\textbf{Setup.} For visual fidelity, we evaluate Fréchet Inception Distance (FID, \cite{heusel2017gans}) of images generated from the random 10K prompts of MS-COCO validation set. For text-alignment and user-preference, we additionally measure ImageReward \citep{xu2024imagereward} and PickScore \citep{kirstain2023pick} which are more reliable compared to the existing text-image scoring metrics, such as CLIP similarity. For Table \ref{table: result}, we use 4 step Euler sampling for SDXL-Lightning and its LoRA variant, 4 step iterative random sampling for DMD2, LCM, and LCM-LoRA (incompatible with conventional solvers), and 6 step DPM++ 2S Ancestral sampling \cite{lu2022dpmpp} for SDXL-Turbo, utilizing DreamShaper \cite{dreamshaper2024}, an open-source customized model from the community. 
\vspace{-0.1cm}

\begin{figure}[htbp]
    \centering
    \includegraphics[width=\linewidth]{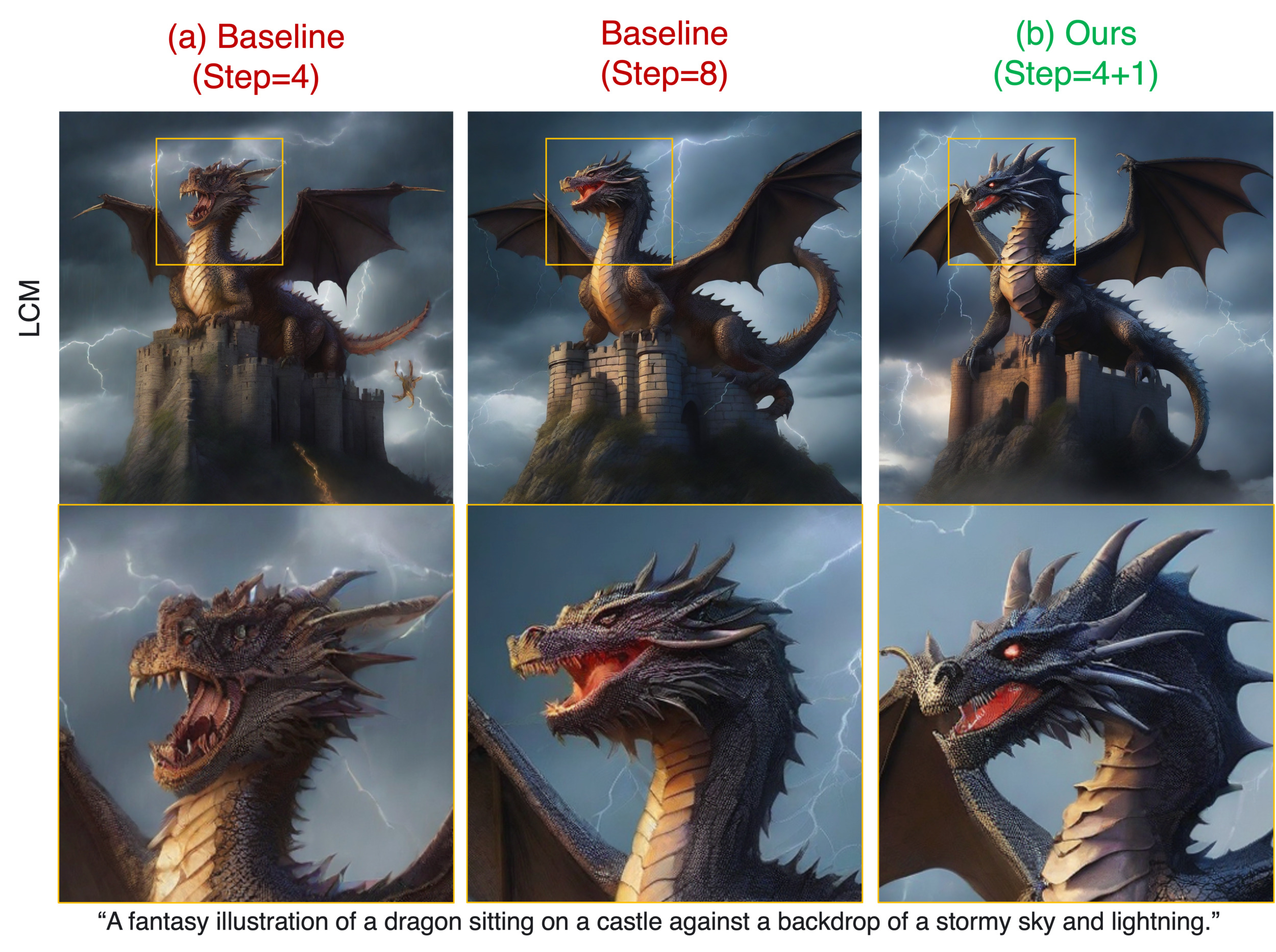}
    \caption{{(\textbf{a}) Results of baseline (LCM) with 4 and 8 sampling steps. (\textbf{b}) Ours with 4 step sampling + 1 step distillation.}}
    \vspace{-0.4cm}
    \label{fig: cost2}
\end{figure}

\subsection{Evaluation and analysis}
\label{sec: result}
\vspace{-0.1cm}

\begin{table}[htbp]
\caption{Quantitative evaluation on MS-COCO 10K with 4 step baseline sampling and 1 step additional inference-time distillation. Light refers to SDXL-Lightning.} 
\vspace{-0.1cm}
\label{table: result}
\centering
\resizebox{\linewidth}{!}{
\begin{tabular}{c c c c}
\toprule
{Models} & {FID ($\downarrow$)} & {ImageReward ($\uparrow$)} & {PickScore ($\uparrow$)} \\
\midrule
{LCM \cite{luo2023latent}} & {20.674} & {0.561} & {0.494} \\
\rowcolor{ppgreen}
{LCM++} & {\textbf{20.149}} & {\textbf{0.597}} & {\textbf{0.505}} \\
\midrule
{LCM-LoRA \cite{luo2023lcmlora}} & {20.300} & {0.494} & {0.490} \\
\rowcolor{ppgreen}
{LCM-LoRA++} & {\textbf{19.815}} & {\textbf{0.522}} & {\textbf{0.510}} \\
\midrule
{Light \cite{lin2024sdxl}} & {24.506} & {0.787} & {0.496} \\
\rowcolor{ppgreen}
{Light++} & {\textbf{23.876}} & {\textbf{0.820}} & {\textbf{0.503}} \\
\midrule
{Light-LoRA \cite{lin2024sdxl}} & {25.304} & {0.750} & {0.482} \\
\rowcolor{ppgreen}
{Light-LoRA++} & {\textbf{24.429}} & {\textbf{0.778}} & {\textbf{0.518}} \\
\midrule
{DMD2 \cite{yin2024improved}} & {21.238} & {0.777} & {0.490} \\
\rowcolor{ppgreen}
{DMD2++} & {\textbf{20.937}} & {\textbf{0.797}} & {\textbf{0.510}} \\
\midrule
{Turbo \cite{sauer2023adversarial}} & {18.612} & {0.296} & {0.499} \\
\rowcolor{ppgreen}
{Turbo++} & {\textbf{18.481}} & {\textbf{0.310}} & {\textbf{0.501}} \\
\bottomrule
\end{tabular}
}
\end{table}

\textbf{Quantitative Analyses.} As shown in Table \ref{table: result}, Distillation++ consistently improves visual quality, text alignment, and user preference across various distillation model baselines and solvers, with only a single additional step of teacher (SDXL) evaluation. Most multi-step distillation models offers only a preset number of sampling steps with a fixed time-step schedule, limiting sampling flexibility. Distillation++ addresses this limitation and provide a post-training option for sampling steps, effectively reduces artifacts, improving fidelity and semantic alignment \cite{park2024energy}, as shown in Fig. \ref{fig: result}.
It is important to note that Distillation++ is a synergistic combination with student and teacher models, as the same number of samplign steps are used with only teacher model (SDXL), the results are inferior with many artifacts and semantic misalignment as demonstrated in Fig.~\ref{fig: teaser} and Supplementary.

\noindent\textbf{Computational Costs and Performance.} Distillation++ serves as an efficient sampling correction, though it introduces additional function evaluation: (1) Since the spatial layout is largely determined in the early sampling stages, early-stage guidance is required to rectify the physical feasibility and text adherence. (2) Increasing the sampling steps of the student may not guarantee these corrections due to the inherent gap between student and teacher models. For instance, Fig. \ref{fig: cost1} and \ref{fig: cost2} demonstrate that while additional steps in student only models may enhance fidelity, it may not be sufficient to capture the intended semantics, e.g. `cat \textit{patting}' or improves physical feasibility, e.g. well-structured spaceship launcher. Distillation++ mitigates this and nudges the sampling path towards a more feasible region.
Table \ref{table: same time} supports these observations where we have compared the wall-clock time and performance of Distillation++ with its counterpart baselines. It shows that the one-step inference-time distillation takes comparable or even shorter wall time compared to that of a single student sampling step, even though the teacher model relies on CFG. This is attributed to the parallel computing of both student and teacher models and batch-wise prediction of conditional and unconditional scores. That said, Distillation++ achieves consistent performance improvement without compromising significant computational costs as shown in Table \ref{table: same time}.

\begin{table}[htbp]
\caption{Comparative study on wall time and generation performance (MS-COCO 10K). $4+1$ step refers to a 4 step student model sampling and 1 step inference-time distillation. Wall-clock time is measured per each prompt.}
\vspace{-0.1cm}
\label{table: same time}
\centering
\resizebox{\linewidth}{!}{
\begin{tabular}{c c c c c c c c c}
\toprule
\multirow{2}{*}{Metrics} & \multicolumn{4}{c}{LCM \cite{luo2023latent}} & \multicolumn{4}{c}{LCM-LoRA \cite{luo2023lcmlora}} \\
\cmidrule(l){2-5} \cmidrule(l){6-9}
{} & {4+1 step} & {5 step} & {5+1 step} & {6 step} & {4+1 step} & {5 step} & {5+1 step} & {6 step} \\ 
\cmidrule(l){1-3} \cmidrule(l){4-5} \cmidrule(l){6-7} \cmidrule(l){8-9}
    {FID ($\downarrow$)} & \ccw{{\textbf{20.149}}} & {20.732} & \ccw{{\textbf{20.369}}} & {21.540} & \ccw{{\textbf{19.815}}} & {20.579} & \ccw{{\textbf{20.244}}} & {21.039}   \\
    {ImageReward ($\uparrow$)} & \ccw{{\textbf{0.597}}} & {0.593} & \ccw{{\textbf{0.603}}} & {0.585} & \ccw{{\textbf{0.522}}} & {0.518} & \ccw{{\textbf{0.528}}} & {0.519} \\
    {Time (sec, $\downarrow$)} & \ccw{{\textbf{1.987}}} & {1.996} & \ccw{{\textbf{2.241}}} & {2.250} & \ccw{{\textbf{1.828}}} & {1.830} & \ccw{{\textbf{2.067}}} & {2.085} \\
\bottomrule
\end{tabular}
}
\end{table}

\noindent\textbf{Ablation study on renoising process.} Table \ref{table: time schedule} highlights the importance of decreasing (reverse-diffusion) time step schedule in the renoising process \eqref{eq: sds loss} of Distillation++, where $s=t-\triangle t$ outperforms random timestep renoising strategy or $s=t$. This is in line with empirical findings from prior score distillation studies \cite{kim2024dreamsampler, zhu2023hifa, wu2024consistent3d}, which contrasts with the random timestep sampling used as standards in prior SDS works \citep{poole2022dreamfusion, lin2023magic3d}. We hypothesize this effect arises primarily from the progressive refinement of estimates and hierarchical minimization of KL-divergence score distillation loss, which warrants further exploration in future SDS works.

\begin{figure}[htbp]
    \centering
    \includegraphics[width=\linewidth]{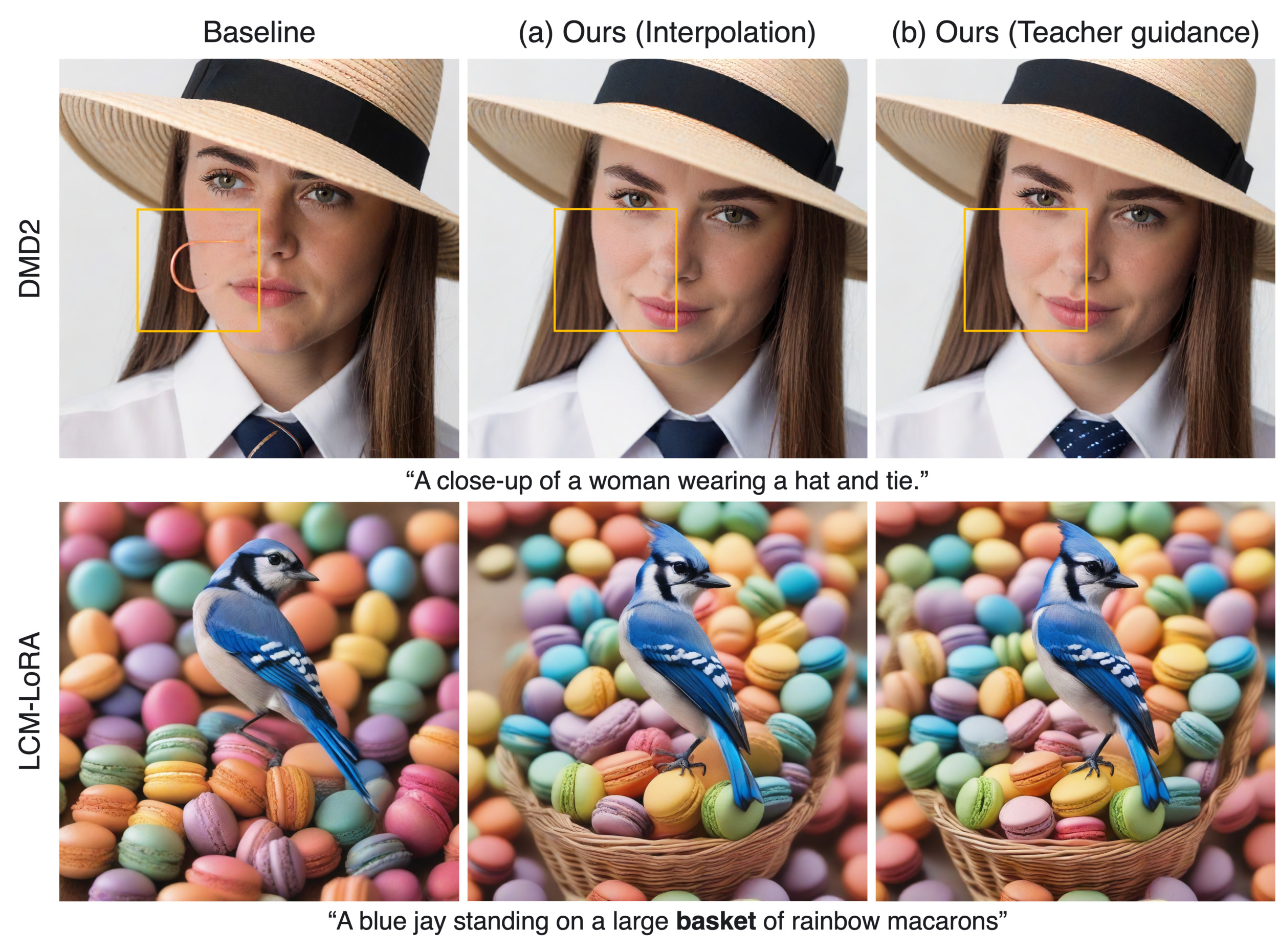}
    \caption{{Comparison on inference-time distillation using (\textbf{a}) interpolative denoising process in \eqref{eq: teacher guidance2}, and (\textbf{b}) teacher guidance \eqref{eq: teacher guidance3}.}}
    \vspace{-0.5cm}
    \label{fig: teacher guidance}
\end{figure}

\noindent\textbf{Teacher guidance.} Figure \ref{fig: teacher guidance} compares Distillation++ with interpolative denoising in \eqref{eq: teacher guidance2} against teacher guided approximation in \eqref{eq: teacher guidance3}. This resemblance shows that the key underlying principle of Distillation++ is to modulate the denoising path under the guidance of teacher models. 



\begin{table}[htbp]
\caption{Ablation study on renoising time schedule (MS-COCO).} 
\label{table: time schedule}
\centering
\resizebox{\linewidth}{!}{
\begin{tabular}{c c c c c}
\toprule
{Models} & {DMD2 \cite{yin2024improved}} & {$s=\text{random} \ t$} & {$s=t$} & {$s=t - \triangle t$} \\
\midrule
{FID ($\downarrow$)} & {21.238} & {21.105} & {21.342} & {\textbf{20.937}} \\
\midrule
{ImageReward ($\uparrow$)} & {0.777} & {0.771} & {0.777} & {\textbf{0.797}} \\
\bottomrule
\end{tabular}
\vspace{-0.9cm}
}
\end{table}

\section{Conclusion}
\vspace{-0.1cm}
{This paper fosters a symbiotic collaboration between two diffusion models: fast but suboptimal student models and slower, high-quality teacher models. Distillation++ serves as a teacher-guided sampling method, minimizing SDS loss by leveraging a pre-trained model to evaluate student estimates during sampling. 

\textbf{Future direction.} This approach may open avenues for exploring diffusion model ensembles or combination with flow matching models \cite{liu2023flow, liu2022flow} for synergistic sampling. Extending Distillation++ to video diffusion distillation would be a promising direction for future work, where video-domain distillation lags behind image-domain quality.}


{
    \small
    \bibliographystyle{ieeenat_fullname}
    \bibliography{main}
}

\clearpage
\setcounter{page}{1}
\maketitlesupplementary

The supplementary sections are organized as follows. 
Section \ref{sec: pseudocode} introduces the pseudo training algorithm behind our inference-time diffusion distillation framework.
In Section \ref{sec: exp setting}, we provide experimental details. Section \ref{sec: additional results} features additional results. Following this, we delve into the future directions and limitations of the proposed method in Section \ref{sec: discussion}. Code will be released in \url{https://github.com/anony-distillationpp/distillation_pp}.
\vspace{-0.1cm}

\section{Pseudo-code}
\vspace{-0.3cm}
\label{sec: pseudocode}
\begin{algorithm}
\label{pseudocode}
	\caption{Inference-time Diffusion model distillation} 
	\begin{algorithmic}[1]
	    \State {\bfseries Input:} Student model \textcolor{student}{$\theta$}, Teacher model \textcolor{teacher}{$\psi$}, $N$ sampling steps, $k$ number of steps of teacher guidance, CFG scale $\omega$, Teacher guidance scale $\lambda$.
        
        \State {\bfseries Output:} Improved generation $\x_0^*$. \\
        \State $\x_T \sim \Nc(\x_T| 0, I), \quad \triangle t = T / N$
        
            \For {$t=T$ \textbf{to} $\triangle t$}
                \State \textbf{Stage 1.} \textit{Initial student estimation}
                \State $\tweediecstudent(t) = \frac{\x_t - \sqrt{1-\alphabar_t} \epshatstudent(\x_t, \c)}{\sqrt{\alphabar_t}}$
                \\
                \State \textbf{Stage 2.} \textit{Revised teacher estimation}
                 \If {$\text{step} < k$}
                    \State Renoising step $s = t - \triangle t$.
                    \State $\x_s = \sqrt{\alphabar_s} \tweediecstudent(t) + \sqrt{1-\alphabar_s} \epsilonb.$ ($\epsilonb \sim \Nc(\epsilonb| 0, I))$
                    \State $\tweediecteacher(s) = \frac{\x_s - \sqrt{1-\alphabar_s} \epshatteacher(\x_s, \c)}{\sqrt{\alphabar_s}}$.
                    \State $\tweedieppc(t) = (1-\lambda) \tweediecstudent(t) + \lambda \tweediecteacher(s)$
                \Else
                    \State $\tweedieppc(t) = \tweediecstudent(t)$
                \EndIf
                \State Update $\x_{t-\triangle t}$ by forwarding $\tweedieppc(t)$. 
            \EndFor
             
	\end{algorithmic} 
\end{algorithm}

We fix typo in \eqref{eq: cfg tweedie} with $\tilde{\x}_t \rightarrow \x_t$. 
Also, we note that $k=1$ used in every quantitative analysis, ensuring computational efficiency.
As the proposed framework revises estimation by interpolation, it can be seamlessly extended with a convex combination of multiple teacher revisions. Moreover, as Algorithm 1 is described with random renoising strategy (Line 12), it is fully compatible with general student models ranging from ones directly predicting the PF-ODE endpoint to the progressive distillation branches. 

\section{Experimental details}
\label{sec: exp setting}
\subsection{Extension to other solvers}
For completeness, we extend Distillation++ to accommodate a broader range of ODE/SDE solvers. The core principle lies in steering the \textit{denoising} process with teacher models. Specifically, we consider solving the variance-exploding (VE) PF-ODE, commonly employed in standard diffusion model implementations\footnote{\url{https://github.com/crowsonkb/k-diffusion}}, which can be readily derived via reparameterization of VP diffusion models. Following the notation in \citet{lu2022dpmpp}, we consider a sequence of timesteps $\{t_i\}_{i=0}^M$, where $t_0 = T$ denotes the initial starting point of the reverse sampling (i.e. Gaussian noise).

\textbf{Euler \cite{karras2022elucidating}.} This is in line with DDIM \cite{song2020denoising} and thus included for completeness:
\begin{align*}
    \x_{t_{i+1}} &= \tweedieppc(\x_{t_i}) + \frac{\x_{t_i} - \tweediecstudent(\x_{t_i})}{\sigma_{t_i}}\cdot \sigma_{t_{i+1}},
    \vspace{-0.1cm}
\end{align*}
where $\tweedieppc(\x_{t_i})$ refers to the revised estimate by interpolation. CFG++ \cite{chung2024cfg++} can be integrated by replacing $\tweediecstudent(\x_{t_i})$ with $\textcolor{student}{\hat{\x}_\varnothing^\theta}(\x_{t_i})$.

\textbf{Euler Ancestral.} 
The Euler Ancestral sampler extends the Euler method by introducing stochasticity, taking larger steps and adding a small random noise. This may potentially improve sampling diversity:
\begin{align*}
    \x_{t_{i+1}} &= \tweedieppc(\x_{t_i}) + \frac{\x_{t_i} - \tweediecstudent(\x_{t_i})}{\sigma_{t_i}}\cdot (\sigma_{t_{d_i}} - \sigma_{t_i}) + \sigma_{t_i} \epsilonb,
\end{align*}
where $t_i > t_{d_i} > t_{i+1}$ and $\epsilonb \sim \Nc(\epsilonb| 0, I)$.

\textbf{DPM-solver++ 2M \cite{lu2022dpmpp}.} While many student models support only first-order solvers, customized distillation models in the open-source community\footnote{\url{https://civitai.com/}} are compatible with higher-order solvers like DPM-Solver++ 2M. Using an iterative process initialized with Gaussian noise, DPM-solver++ refines the sampling trajectory with higher-order corrections, enabling precise updates. Similarly as DPM-solver++ 2S, define $\sigma_t := e^{-t},\,h_i := t_i - t_{i-1},$ and $r_i := h_{i-1}/h_{i}$. Given $\x_{t_0}$ initialized as Gaussian noise, the first iteration reads:
\begin{align*}
    \x_{t_1} &= \tweediecstudent(\x_{t_0}) + e^{-h_1}(\x_{t_0} - \tweediecstudent(\x_{t_0})). 
\end{align*}
Then, the following provides higher-order correction:
\begin{align}
\label{eq:dpmpp_2m_cfg_1}
    \bm{D}_i &= \tweediecstudent(\x_{t_{i-1}}) + \frac{1}{2r_i}\left(
    \tweediecstudent(\x_{t_{i-1}}) - \tweediecstudent(\x_{t_{i-2}})
    \right),\\
    \x_{t_i} &= e^{-h_i}\x_{t_{i-1}} - (e^{-h_i} - 1)\bm{D}_i.
\label{eq:dpmpp_2m_cfg_2}
\end{align}
Rearranging \eqref{eq:dpmpp_2m_cfg_1}, \eqref{eq:dpmpp_2m_cfg_2}, we can rewrite the update steps as
\begin{align}
\label{eq:dpmpp_2m_cfg_rearrange}
    \x_{t_i} &= \tweediecstudent(\x_{t_{i-1}}) - e^{-h_i}\tweediecstudent(\x_{t_{i-1}}) \nonumber \\
    &+ \frac{1 - e^{-h_i}}{2r_i}\left(
    \tweediecstudent(\x_{t_{i-1}}) - \tweediecstudent(\x_{t_{i-2}})
    \right) + e^{-h_i}\x_{t_{i-1}}. \nonumber
\end{align}
As we are interested in modulating the final form of denoised estimates, Distillation++ can be applied as follows:
\begin{align}
    \x_{t_i} &= \tweedieppc(\x_{t_{i-1}}) - e^{-h_i}\tweediecstudent(\x_{t_{i-1}}) \nonumber \\
    &+ \frac{1 - e^{-h_i}}{2r_i}\left(
    \tweediecstudent(\x_{t_{i-1}}) - \tweediecstudent(\x_{t_{i-2}})
    \right) + e^{-h_i}\x_{t_{i-1}}. \nonumber
\end{align}

Ancestral variants (DPM-solver++ 2M A) can be readily derived by similarly adding a random noise.

\subsection{Experiment Setup}

In this work, we employ several diffusion distillation models: DMD2 \cite{yin2024improved}, SDXL-Turbo \cite{sauer2023adversarial}, SDXL-Lightning \cite{lin2024sdxl}, LCM \cite{luo2023latent}, LCM-LoRA \cite{luo2023lcmlora}, and SDXL-Lightning LoRA. These models rely on classifier-free guidance (CFG) with a fixed guidance scale during training. We use $w=7.5$ for CFG with teacher models $\tweediecteacher(s)$.

Baselines using 4 sampling steps include SDXL-Lightning, DMD2, and SDXL-Turbo, while LCM and LCM-LoRA use 8 sampling steps. Specifically, for Table \ref{table: result}, we use 4 step Euler sampling for SDXL-Lightning and its LoRA variant, 4 step iterative random sampling for DMD2, LCM, and LCM-LoRA (incompatible with conventional solvers), and 4 step DPM++ 2S Ancestral sampling \cite{lu2022dpmpp} for SDXL-Turbo, utilizing DreamShaper \cite{dreamshaper2024}, an open-source customized model from the community. 

All quantitative results presented in the main paper are obtained using Algorithm 1, which employs a fully random renoising process (Line 12) to ensure generality. For distillation models that directly approximate the score function (e.g., SDXL-Lightning, SDXL-Turbo), the renoising strategy can be extended by incorporating the predicted epsilon from the previous time step. While the performance gains are marginal, we observed some improvement, likely due to adherence to the fundamental refinement principle of reverse diffusion sampling, as studied in \cite{kim2024dreamsampler}.

For all quantitative analyses, we fix the number of teacher guidance steps at $k=1$. Our approach remains simple with minimal hyperparameters, where the teacher guidance scale 
$\lambda$ is the primary parameter. Specifically, we set 
$\lambda=0.02$ for LCM, LCM-LoRA, and DMD2, and 
$\lambda=0.1$ for SDXL-Turbo, SDXL-Lightning, and SDXL-Lightning LoRA. The same $\lambda$ values are used in Fig.~\ref{fig: teacher guidance} to evaluate the teacher guidance approximation.
%

\begin{figure*}[t!]
    \centering
    \includegraphics[width=0.94\textwidth]{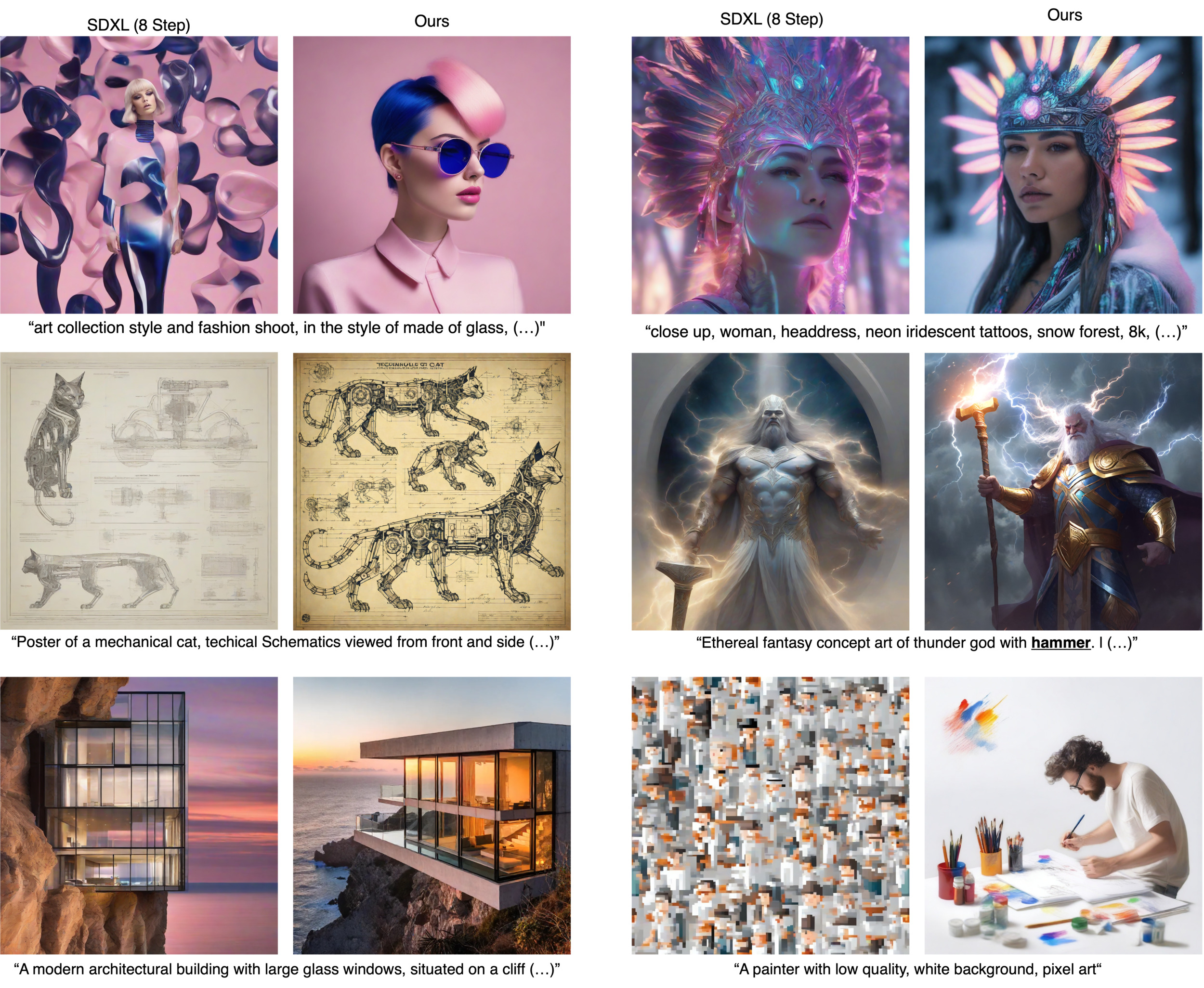}
    \caption{Comparisons between SDXL teacher model (8 steps) and ours. Our results are from the Fig.~\ref{fig: result} of the main paper.}
    \label{fig: teacher compare}
\end{figure*}

\section{Additional results}
\label{sec: additional results}
\begin{figure*}[t!]
    \centering
    \includegraphics[width=\textwidth]{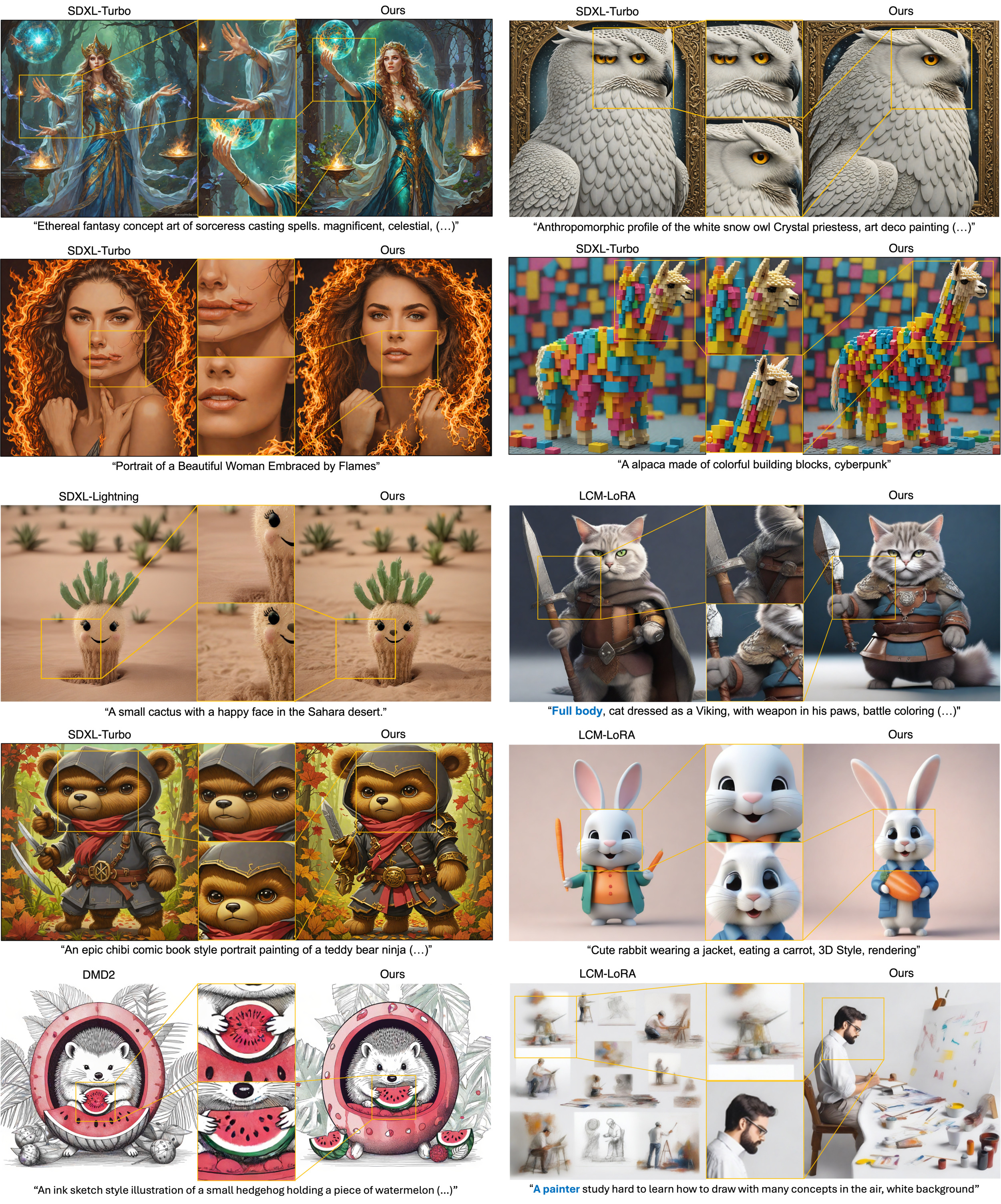}
    \caption{Qualitative comparisons against state-of-the-art distillation baselines. Baselines using 4 sampling steps: SDXL-Lightning, DMD2, SDXL-Turbo. Baselines using 8 sampling steps: LCM, LCM-LoRA.}
    \label{fig: supple results 1}
\end{figure*}

\begin{figure*}[t!]
    \centering
    \includegraphics[width=\textwidth]{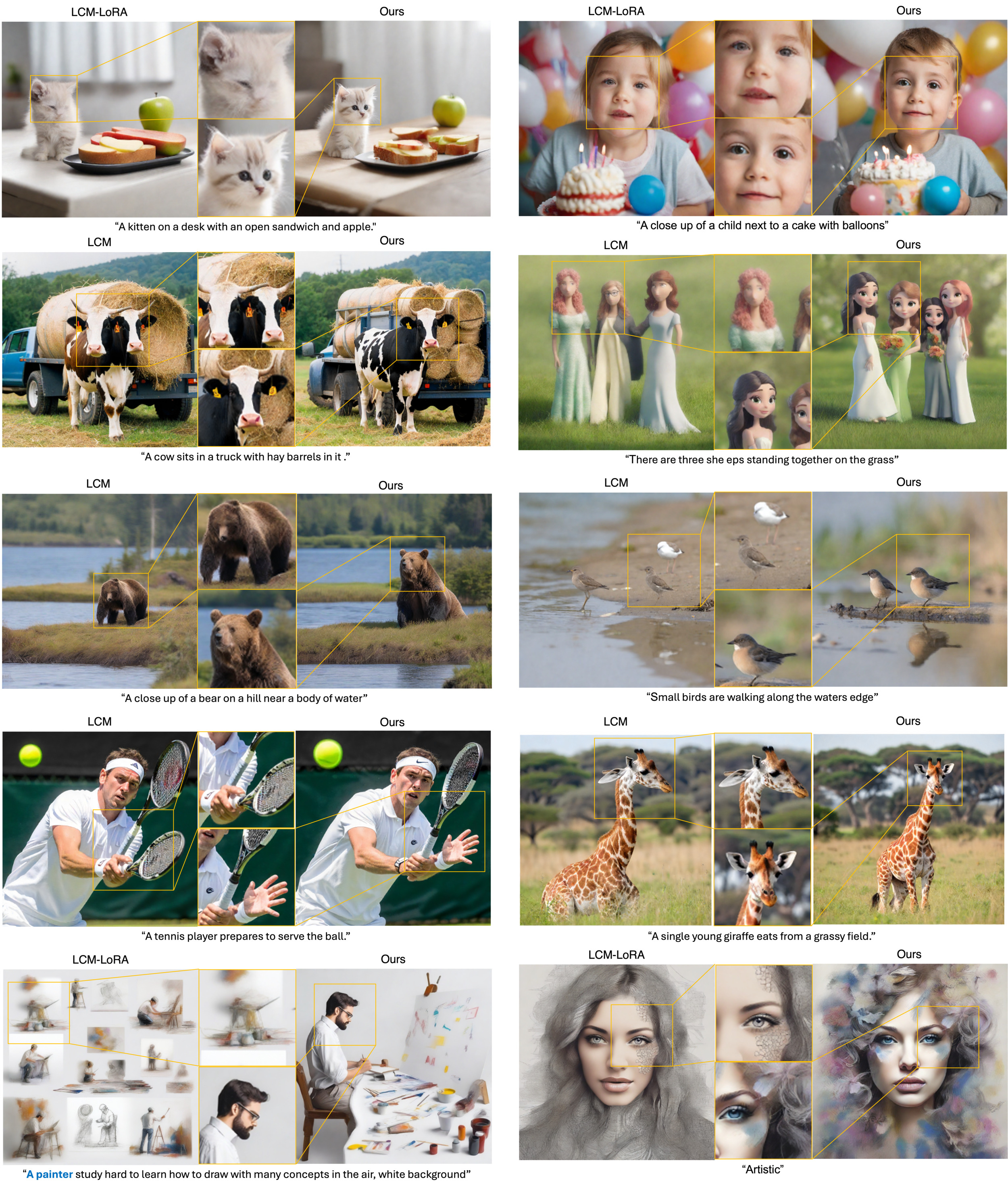}
    \caption{Additional qualitative comparisons against state-of-the-art distillation baselines.}
    \label{fig: supple results 2}
\end{figure*}

In Fig.~\ref{fig: supple results 1} and \ref{fig: supple results 2}, we demonstrate the effectiveness of the proposed inference-time distillation with various student models. This advances stem from the guidance of teacher model, whereas the teacher model itself does not guarantee high-quality samples with few sampling steps, e.g. 8 steps (Fig.~\ref{fig: teacher compare}). That said, our work fosters a synergistic collaboration between two kinds of diffusion models: fast but sub-optimal student models, and high-quality buy computationally expensive teacher models.


\section{Discussions and Limitations}
\label{sec: discussion}
Beyond the image domain, diffusion models have become a cornerstone of high-dimensional visual generative modeling, including applications such as video generation \cite{ho2022video} and multi-view synthesis \cite{voleti2025sv3d}. While computational efficiency is critical for modeling in these high-dimensional spaces, recent studies highlight the challenge of reducing inference steps for video generation. Compared to the image generation, the quality and prompt alignment of generated motion are more dependent on the number of inference steps \cite{polyak2024movie}. Although an increasing number of video diffusion distillation models \cite{zhai2024motion, lin2024animatediff} have emerged, a significant gap remains between student and teacher video diffusion models. Applying inference-time diffusion distillation to the video domain offers a promising avenue for improving temporal consistency, addressing issues that text-to-video (T2V) models--often adapted from text-to-image (T2I) models--frequently encounter.

Additionally, flow-based generative models generalize diffusion models and similarly rely on off-the-shelf ODE solvers. Thus, extending inference-time distillation to bridge the gap between flow-based teacher models (e.g., \cite{esser2024scaling}) and student models (e.g., \cite{sauer2024fast}) presents an intriguing direction for future research.

One limitation of the proposed framework is that both student and teacher (latent) diffusion models must operate within a shared latent space to enable interpolation between denoised estimates. To address this, one potential approach could involve mapping latent estimates back to pixel space, refining the student’s pixel estimates, and subsequently re-encoding them. Furthermore, the interplay between students and diverse open-source customized teacher models, which exhibit varying styles and aesthetics, represents another compelling avenue for exploration.

\end{document}